\documentclass[pdflatex,sn-apa,iicol]{sn-jnl}

\usepackage{graphicx}%
\usepackage{multirow}%
\usepackage{amsmath,amssymb,amsfonts}%
\usepackage{amsthm}%
\usepackage{mathrsfs}%
\usepackage[title]{appendix}%
\usepackage{xcolor}%
\usepackage{textcomp}%
\usepackage{manyfoot}%
\usepackage{booktabs}%
\usepackage{algorithm}%
\usepackage{algorithmicx}%
\usepackage{algpseudocode}%
\usepackage{listings}%
\usepackage{amssymb}%
\usepackage{tikz}%

\usepackage[capitalize]{cleveref}%
\crefname{section}{Sec.}{Secs.}%
\Crefname{section}{Section}{Sections}%
\crefname{figure}{Fig.}{Figs.}%
\Crefname{figure}{Figure}{Figures}%

\theoremstyle{thmstyleone}%
\theoremstyle{thmstyletwo}%

\theoremstyle{thmstylethree}%

\def\modelName{SAGANet}
\def\datasetName{Segmented Music Solos}

\newcommand{\tikzxmark}{%
\tikz[scale=0.23] {
    \draw[line width=0.7,line cap=round] (0,0) to [bend left=6] (1,1);
    \draw[line width=0.7,line cap=round] (0.2,0.95) to [bend right=3] (0.8,0.05);
}}
\newcommand{\tikzcmark}{%
\tikz[scale=0.23] {
    \draw[line width=0.7,line cap=round] (0.25,0) to [bend left=10] (1,1);
    \draw[line width=0.8,line cap=round] (0,0.35) to [bend right=1] (0.23,0);
}}

\begin{document}

\title[Video Object Segmentation-aware Audio Generation]{Video Object Segmentation-aware Audio Generation}

\author[1]{\fnm{Ilpo} \sur{Viertola}}\email{ilpo.viertola@tuni.fi}

\author[2]{\fnm{Vladimir} \sur{Iashin}}\email{vi@robots.ox.ac.uk}

\author[1]{\fnm{Esa} \sur{Rahtu}}\email{esa.rahtu@tuni.fi}

\affil[1]{\orgdiv{Computer Vision Group}, \orgname{Tampere University}, \orgaddress{\street{Korkeakoulunkatu 7}, \city{Tampere}, \postcode{33720}, \country{Finland}}}

\affil[2]{\orgdiv{Visual Geometry Group}, \orgname{University of Oxford}, \orgaddress{\street{25 Banbury Rd}, \city{Oxford}, \postcode{OX2 6NN}, \country{United Kingdom}}}


\abstract{Existing multimodal audio generation models often lack precise user control, which limits their applicability in professional Foley workflows.
In particular, these models focus on the entire video and do not provide precise methods for prioritizing a specific object within a scene, generating unnecessary background sounds, or focusing on the wrong objects.
To address this gap, we introduce the novel task of video object segmentation-aware audio generation, which explicitly conditions sound synthesis on object-level segmentation maps.
We present \modelName, a new multimodal generative model that enables controllable audio generation for musical instruments by leveraging visual segmentation masks along with video and textual cues.
Our model provides users with fine-grained and visually localized control over audio generation.
To support this task and further research on segmentation-aware Foley, we propose \datasetName, a benchmark dataset of musical instrument performance videos with segmentation information.
Our method demonstrates substantial improvements over current state-of-the-art methods and sets a new standard for controllable, high-fidelity Foley synthesis for musical audio.
Code, samples, and \datasetName\space are available at \textbf{\url{saganet.notion.site}}.}


\maketitle

\section{Introduction}\label{sec1}

Multimodal audio generation focuses on synthesizing audio, given a conditional video feed, a textual description, or conditions in other modalities.
These models can be utilized in \textit{Foley} processing, where the goal is to produce a soundtrack for a video.
For the artificial Foley models to thrive, \textit{e.g.}, as a video post-processing tool, the end-user has to have high control over the synthesized result.
In addition to controllability, the model has to produce high-quality samples with great semantic and temporal accuracy.

Recent advances in diffusion \citep{liuAudioLDMTexttoAudioGeneration2023} and conditional flow matching (CFM) \citep{tongImprovingGeneralizingFlowbased2024} models have improved the audio quality of the artificial Foley models \citep{chengTamingMultimodalJoint2024, chenVideoGuidedFoleySound2024, moTexttoAudioGenerationSynchronized2024, wangFrierenEfficientVideotoAudio2024, xuVideotoAudioGenerationHidden2024, wangV2AMapperLightweightSolution2023, polyakMovieGenCast2025}.
However, existing models still lack precise user guidance and control.
Although recent models \citep{chengTamingMultimodalJoint2024, chenVideoGuidedFoleySound2024, polyakMovieGenCast2025, moTexttoAudioGenerationSynchronized2024} introduce more modalities to guide the audio generation process, they still lack precision.
In complex scenes, describing the target object using \textit{e.g.}, text, quickly becomes unfeasible.

To improve control in Foley generation, we propose a new audio synthesis task: video object segmentation-aware audio generation.
This task focuses on nuanced control, highlighting the model's ability to generate audio for a specific object in the video rather than for the full scene.
To the best of our knowledge, we are the first to utilise visual segmentation information in the audio generation task.

Existing state-of-the-art methods \citep{chengTamingMultimodalJoint2024, chenVideoGuidedFoleySound2024, viertolaTemporallyAlignedAudio2024, wangFrierenEfficientVideotoAudio2024, polyakMovieGenCast2025} focus on training the models from scratch, utilizing large uni- and multimodal datasets.
It demands significant computational resources and quickly becomes infeasible in academic environments.
For example, Movie Gen Audio \citep{polyakMovieGenCast2025} was pretrained on 384 high-end H100 GPUs for 14 days and later fine-tuned on 64 H100s for 24 hours.
MMAudio \citep{chengTamingMultimodalJoint2024} was trained on the same high-end GPU hardware, only utilizing fewer GPU hours per training run.

Alternatively, recent works reduce training costs by using lightweight \textit{aligners} or \textit{adapters} to condition pretrained text-to-audio models on video sequences \citep{wangV2AMapperLightweightSolution2023,moTexttoAudioGenerationSynchronized2024,jeongReadWatchScream2024,xingSeeingHearingOpendomain2024}.
While these models are more efficient to train, they often struggle with temporal alignment, as fine-tuning text-to-audio models without built-in temporal control remains challenging.
The concurrent work of \cite{li2025sounding} also explores object-aware audio generation.
However, their focus on image-to-audio generation makes the approach unusable for artificial Foley.

Our approach draws on recent advances in multimodal audio generation \citep{chengTamingMultimodalJoint2024} and localized image captioning \citep{lian2025describe}.
To enable segmentation-aware audio generation for musical instruments, we develop a self-supervised control module on top of a pretrained network that enables users to select a specific object in a video to generate sound for.
By training our control component on a small yet high-quality dataset, we achieve better controllability, temporal synchronization, and semantic quality compared to the original model, training only a fraction of the parameters.
Although our model is trained on videos with a single audio source, it can generalize to scenes with multiple audio sources and generate audio for the target object.

To train the segmentation-aware control module, we propose a high-quality dataset with sounding object segmentation maps and a high audio-video correspondence.
The dataset consists of solo acts played on a variety of musical instruments.
We curate the dataset based on Solos \citep{montesinos2020solos}, AVSBench \citep{zhou2022audio, zhou2023audio}, and MUSIC21 \citep{zhao2018sound, zhao2019sound} datasets.
We design a pipeline that generates short video clips given the original videos, ensuring that the target sounding object is present in both the auditory and visual modalities.
Additionally, we extract visual segmentation maps of the target (sounding) object.
We refer to this dataset as \datasetName.
For testing purposes, we utilize the University of Rochester Multi-Modal Music Performance (URMP) dataset \citep{li2018creating}.

Our contributions can be summarized as follows: i) a new audio synthesis task, namely video object segmentation-aware audio generation, ii) a video object segmentation-aware control for a state-of-the-art multimodal generative audio model, iii) we show that by training our model with single-source samples it can generate audio for target object in multi-source scenes, and iv) a new benchmark dataset, \datasetName, with sounding object segmentation information.
This manuscript is based on a previous conference publication at German Conference on Pattern Recognition (GCPR) 2025, which we significantly consolidate here by elaborating on evaluation methods, adding ablation studies and subjective results, and conduncting a subjective evaluation.

\section{Related Work}

\subsection{Artificial Foley Models}
\label{ssec:gen-audio-models}
Artificial Foley models have gained considerable popularity.
Many published models are built on top of autoregressive transformer architecture \citep{viertolaTemporallyAlignedAudio2024, meiFoleyGenVisuallyGuidedAudio2023, shefferHearYourTrue2023, iashinTamingVisuallyGuided2021}.
Also, another transformer-based approach is to utilize Masked Generative Image Transformer (MaskGIT) \citep{changMaskGITMaskedGenerative2022} schema for the audio generation task \citep{zivMaskedAudioGeneration2024, pascualMaskedGenerativeVideotoAudio2024, liuTellWhatYou2024, suVisionAudioUnified2024}.
Another popular approach is to utilize diffusion \citep{xuVideotoAudioGenerationHidden2024,luoDiffFoleySynchronizedVideotoAudio2023,chenAction2SoundAmbientAwareGeneration2024} or flow-matching methods \citep{wangFrierenEfficientVideotoAudio2024, chengTamingMultimodalJoint2024}.
To avoid resource- and time-exhaustive training from scratch, prior work has explored using pretrained text-to-audio models for video-to-audio generation by training lightweight feature aligners or control modules between the modalities \citep{wangV2AMapperLightweightSolution2023, moTexttoAudioGenerationSynchronized2024, jeongReadWatchScream2024, xingSeeingHearingOpendomain2024}.
Although these models achieve good audio quality, they often struggle to generate temporally aligned audio.
When the text modality is fixed during training, learning an aligned feature space between modalities becomes challenging.

The current Foley methods lack controllability.
For artists to fulfil their needs in applications such as video post-processing, they must have fine-grained control over the model.
One way to add control is to introduce conditioning signals from other modalities, \textit{e.g.}, text.

Although early multimodal approaches, \textit{e.g.} text-and-video-to-audio models, did not meet the generation quality compared with dedicated video-to-audio models \citep{ruanMMDiffusionLearningMultiModal2023,kimVersatileDiffusionTransformer2024, tangAnytoAnyGenerationComposable2023,tangCoDi2InContextInterleaved2023}, recent work \citep{chengTamingMultimodalJoint2024,chenVideoGuidedFoleySound2024,polyakMovieGenCast2025} has shown that artificial Foley models can benefit from multimodality.
For example, \cite{chengTamingMultimodalJoint2024} show that training a generative audio network with combined text-audio and text-video-audio data enables high generation quality while preserving temporal and semantic alignment with video and text.
However, the user controllability remains limited since describing the target objects with textual prompts in complex scenes can be difficult or infeasible.
To tackle this, we develop a novel model that supports audio generation conditioned with text and video, but also allows users to define the sounding object with a semantic mask.
In practice, user can click an object from a video frame, and our model generates audio for that specific object.

\subsection{Video Object Segmentation}
Video Object Segmentation (VOS) refers to the task of segmenting and tracking objects at the pixel level across video frames while maintaining temporal consistency \citep{pont-tuset2017DAVISChallenge2018}.
It typically involves distinguishing foreground objects, such as people, animals, or vehicles, from the background.

Different types of VOS tasks include semi-supervised and unsupervised VOS.
In semi-supervised VOS, a ground-truth mask is provided for the target object in the first frame, and the goal is to segment the object in the remaining frames \citep{pont-tuset2017DAVISChallenge2018}.
In unsupervised VOS, no initial mask is given.
The goal is to discover and segment prominent objects automatically \citep{fragkiadaki2015learning}.

Recent advancements have introduced promptable VOS models.
SAM2 \citep{raviSAM2Segment2024} is a widely adopted and powerful semi-supervised video object segmentation model capable of real-time object segmentation with images and videos.
The initial mask of the segmented object can be provided manually or based on location coordinates.
SAM2 facilitates applications such as video editing, mixed reality experiences, and efficient annotation of visual data for training computer vision systems.

GroundedSAM2 \citep{ren2024grounded} combines SAM2 with grounding models in a single pipeline, enabling grounding and tracking anything in videos.
GroundedSAM2 enables segmenting objects based on natural language queries, making it a powerful tool, \textit{e.g.}, in data generation.
We utilize GroundedSAM2 with Florence-2 \citep{xiao2023florence} foundation model by prompting it with text labels of target objects and SAM2 by prompting it with location coordinates in our data generation pipeline.
The resulting segmentation maps of sounding objects are then used during training.

\begin{figure*}[t]
    \centering
    \includegraphics[width=0.85\linewidth]{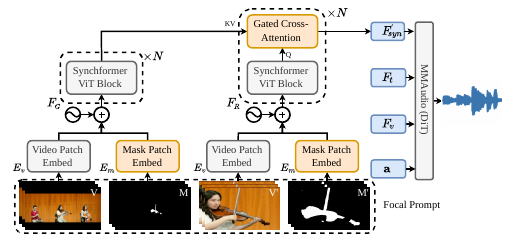}
    \caption{Overview of \modelName~control module. Given a video and its corresponding segmentation masks, the model combines global and local information streams. Gated Cross-Attention layers \citep{alayrac2022flamingo, li2022blip} are used to fuse global and local features extracted by Synchformer \citep{iashin2024synchformer}, with shared weights across both branches. Only the layers highlighted in orange are updated during training. The input features are highlighted with blue colour. $K$ stands for keys, $V$ for values, and $Q$ for queries on cross-attention calculation. $N$ represents layer count. The final audio is generated following the same procedure as in the base MMAudio model. For additional details on MMAudio, refer to work by \cite{chengTamingMultimodalJoint2024}.}
    \label{fig:arc}
\end{figure*}

\subsection{Controlled Generation with Pretrained Audio Networks}
Adapting large pretrained networks to new conditioning inputs is a widely studied topic.
In the diffusion model domain, training a ControlNet \citep{zhangAddingConditionalControl2023} is a popular approach \citep{moTexttoAudioGenerationSynchronized2024, jeongReadWatchScream2024, wuMusicControlNetMultiple2023, houEditingMusicMelody2025}.
ControlNet enables fine-grained control by injecting conditioning features, such as visual cues, semantic tags, or motion information, into the denoising process of a frozen diffusion model.

However, we hypothesize that training a parallel ControlNet \citep{zhangAddingConditionalControl2023} to condition MMAudio \citep{chengTamingMultimodalJoint2024} with video object segmentation information is unnecessary.
\cite{lian2025describe} show that in visual captioning, combining local and global visual features from the same extractor using learnable gated cross-attention \citep{alayrac2022flamingo, li2022blip}, along with object segmentation masks, improves localized captions.
Their large transformer-based \citep{vaswaniAttentionAllYou2023} captioning model is kept frozen, and only the feature extraction process is modified.
Motivated by their findings, we design and implement a localized visual feature extraction model in parallel to MMAudio's \citep{chengTamingMultimodalJoint2024} Synchformer-based \citep{iashin2024synchformer} global feature extractor.

Introducing the segmentation information already at the feature extraction stage allows us to use the same control module across all variants of the generative model.
Thus, the number of trainable parameters remains the same even though the generative network size increases.
In contrast, fusing the segmentation information via a ControlNet-based approach would require training a separate control module per model variant.
Our approach integrates visual segmentation masks as a new conditioning modality, without requiring full model fine-tuning or training of variant-specific control modules.

\section{Method}
We introduce a novel video object segmentation-aware audio generation task and propose a generative audio network that offers fine-grained user control through multiple input modalities: text, video, and video object segmentation masks.
Our approach builds on top of the MMAudio model \citep{chengTamingMultimodalJoint2024}, a state-of-the-art audio generation model conditioned on text and video.
We extend MMAudio with a segmentation-aware control module, resulting in the proposed model, Segmentation-Aware Generative Audio Network (\modelName).
To the best of our knowledge, we are the first to utilize visual segmentation information in the audio generation task.
In addition to the improved usability of the original model, segmentation information enhances the quality and alignment of the synthesized audio.
The overall architecture is shown in \cref{fig:arc}.

\subsection{Preliminary: MMAudio}
\label{ssec:mmaudio}

MMAudio is a conditional flow matching (CFM) based \citep{liuFlowStraightFast2022a, lipmanFlowMatchingGenerative2023, tongImprovingGeneralizingFlowbased2024, wangFrierenEfficientVideotoAudio2024} generative audio model, utilizing Diffusion Transformer (DiT) \citep{peeblesScalableDiffusionModels2023} to approximate the velocity vector field.
DiT's sequence modelling capabilities and range of conditioning methods suit a multimodal setting, where signals are combined across textual, auditory, and visual modalities.
Since MMAudio combines textual, auditory, and visual data, it utilizes a vast amount of different feature extractors \citep{chengTamingMultimodalJoint2024}.

\subsubsection{Conditional Flow Matching}
\label{ssec:cfm}
Conditional flow matching (CFM) \citep{liuFlowStraightFast2022a, lipmanFlowMatchingGenerative2023, tongImprovingGeneralizingFlowbased2024, wangFrierenEfficientVideotoAudio2024} views the conditional generation problem as a conditional mapping from a noise distribution $\mathbf{x}_0\sim p_0(\mathbf{x})$ to data distribution $\mathbf{x}_1\sim p_1(\mathbf{x})$ (\textit{i.e.}, compressed mel spectrogram latent).
It is a time-dependent process of updating the probability density $p_t$ defined by the ordinary differential equation (ODE):
\begin{equation}
   d\mathbf{x}=\mathbf{u}(\mathbf{x}_t|\mathbf{c})dt,\space t\in[0,1],
\end{equation}
where $t$ represents the time, $\mathbf{x}$ is probability density space at time $t$, $\mathbf{u}$ is the value of the velocity vector field (\textit{i.e.}, a gradient field of the probability density map $p_t$ w.r.t $t$) at $\mathbf{x}$, and $\mathbf{c}$ is the conditioning (\textit{i.e.}, video, text, and mask features).
In flow matching, we utilize a deep neural network parametrized by $\theta$ ($\mathbf{v}_\theta$) to represent the velocity vector field $\mathbf{u}$.
Thus, the flow matching objective becomes:
\begin{equation}
\begin{split}
   \mathcal{L}_{\text{FM}}(\theta) = \mathbb{E}_{t, p_t(\mathbf{x})} \| & \mathbf{v}_\theta(\mathbf{x}_t \mid \mathbf{c}; \theta) \\ & - \mathbf{u}(\mathbf{x}_t \mid \mathbf{c}) \|^2,
\end{split}
\label{eq:fm}
\end{equation}
where $p_t(\mathbf{x})$ is the probability distribution of $\mathbf{x}$ at time $t$.
However, because the knowledge of the original data distribution $p_1(\mathbf{x})$ and forms of $p_t$ and $\mathbf{u}$ are limited, computing $\mathbf{u}(\mathbf{x}_t|\mathbf{c})$ becomes infeasible.
Thus, the conditional flow matching objective is derived as follows:
\begin{equation}
\begin{split}
   \mathcal{L}_{\text{CFM}}(\theta) = \mathbb{E}_{t, p_1(\mathbf{x}_1), p_t(\mathbf{x} \mid \mathbf{x}_1)} \| & \mathbf{v}(\mathbf{x}_t \mid \mathbf{c}; \theta) \\ - & \mathbf{u}(\mathbf{x}_t \mid \mathbf{x}_1, \mathbf{c}) \|^2.
\end{split}
\label{eq:cfm}
\end{equation}

\cite{lipmanFlowMatchingGenerative2023} show that \cref{eq:cfm} has identical gradients with \cref{eq:fm}.
For each training step of the vector field estimator, we sample the data point $\mathbf{x}_1$ and noise point $\mathbf{x}_0$ from $p_1(\mathbf{x})$ and $p_0(\mathbf{x})$, respectively.
The network is then optimized using the rectified flow matching (RFM) loss:
\begin{equation}
\left\| v(\mathbf{x}_t \mid \mathbf{c}; \theta) - (\mathbf{x}_1 - \mathbf{x}_0) \right\|^2.
\end{equation}

CFM aims to create straight paths between noise and data by defining a linear interpolation schedule $\mathbf{x}_t=t\mathbf{x}_1-(1-t)\mathbf{x}_0$, where $t\in[0,1]$.
Once the vector estimator network has been trained, different solvers can be used to approximate the solution of the ODE $d\mathbf{x} = v(\mathbf{x}_t \mid \mathbf{c}; \theta)$ at discrete time steps.
A commonly used ODE solver is the Euler method:
\begin{equation}
\mathbf{x}_{t+\epsilon} = \mathbf{x} + \epsilon v(\mathbf{x}_t \mid \mathbf{c}; \theta),
\end{equation}
where $\epsilon$ represents the step size. For more details, please refer to \citep{lipmanFlowMatchingGenerative2023, tongImprovingGeneralizingFlowbased2024, liuFlowStraightFast2022a}.

\subsubsection{Visual Embeddings}
For visual data, MMAudio combines features extracted with CLIP \citep{radford2021learning} and with Synchformer's AVCLIP \citep{iashin2024synchformer} models.
The visual features extracted with CLIP act as global visual embeddings, providing the model with the semantic context of the visual feed.
Synchformer features are used as synchronization cues, providing the model with accurate temporal information.
We design our segmentation-aware control branch parallel to the Synchformer branch.
In contrast to global visual CLIP features, Synchformer captures fine-grained visual features relevant to audio events since it is trained to capture misalignment between audio and video.

\subsubsection{Text Embeddings}
To encode text, MMAudio utilizes CLIP's \citep{radford2021learning} text encoder.
Text serves as an intuitive conditioning input for audio synthesis and also enables the use of high-quality text–audio datasets.
The quality of existing video-audio datasets is sub-optimal, since they are often noisy and consist of samples scraped from platforms such as YouTube \citep{chen2020vggsound, gemmeke2017audio}.
Large manually annotated audio-text datasets, such as AudioCaps \citep{kim2019audiocaps} and WavCaps \citep{mei2024wavcaps}, serve as high-quality training data for MMAudio.

\subsubsection{Audio Embeddings}
The generative process of MMAudio operates in the latent space.
Raw waveforms are represented as magnitude components of Mel spectrograms \citep{stevens1937scale}.
Using a pretrained variational autoencoder \citep{kingma2013auto}, the spectrograms are encoded into latents.
A pretrained vocoder decodes the latent representation into a waveform representation \citep{lee2022bigvgan}.

\subsubsection{Diffusion Transformer}
MMAudio \citep{chengTamingMultimodalJoint2024} introduces a novel multimodal DiT \citep{peeblesScalableDiffusionModels2023} network which synthesises audio given visual and textual conditions.
The blocks of the MMAudio DiT can be split into two different types, \textit{joint} (\textit{i.e.} multimodal) \citep{esserScalingRectifiedFlow2024} and \textit{fused} (\textit{i.e.} unimodal) \citep{flux2024} attention blocks.
Joint attention blocks aim to incorporate information across the modalities.
Following \citep{esserScalingRectifiedFlow2024}, MMAudio concatenates query, key, and value representations across the different modalities (\textit{i.e.}, video, text, and audio) and applies a single joint attention operation \citep{vaswaniAttentionAllYou2023}.
\cite{chengTamingMultimodalJoint2024} also introduce a conditional synchronization module.
It is a token-level conditioning, fusing temporally matched Synchformer features with audio latents using AdaLN \citep{peeblesScalableDiffusionModels2023} layers.
We implement our segment-aware control branch into this conditional synchronization module.

Joint attention blocks are followed by fused attention blocks, which process only auditory information along with global conditioning.
Global conditioning infuses global features (\textit{i.e.}, average-pooled visual and average-pooled textual features) through scales and biases in adaptive layer normalization \citep{peeblesScalableDiffusionModels2023} layers.
Auditory tokens in fused blocks capture information across all modalities by going through joint attention blocks before entering the audio-only fused blocks.

\subsection{Video Object Segmentation-Aware Audio Generation Task Formulation}
\label{ssec:task}
To challenge the precise control in artificial Foley models, we introduce a new audio synthesis task: video object segmentation-aware audio generation.
To support the proposed task, we curate a dataset tailored to it (\cref{sec:dataset}).
In VOS-aware audio generation, the goal is to generate audio focused on the specified region within the video.
Current artificial Foley models focus on the full scene, lacking fine-grained user controllability.
However, focusing only on the small region is not enough, since the overall context is still vital in the generation of credible audio.
For example, the perceived sound of a car is different in a parking hall and on an open road.

Formally, given a visual stream $V \in \mathbb{R}^{T_v \times H \times W \times 3}$, ($T_v$ is frame count, $H$ is height, $W$ is width, and 3 is RGB color channels), and corresponding stream of binary masks $M \in \{0, 1 \}^{T_v \times H \times W \times 1}$, the goal is to produce audio focusing on the specified region in the visual stream $A = G_M(V, M, ...) \in \mathbb{R}^{T_a}$, ($G_M$ is the generative model and $T_a$ is a temporal dimension).
Input modalities are model-dependent, and different models might leverage additional ones.

\subsection{\modelName}
\label{ssec:arc}
Drawing on the success of DAM \citep{lian2025describe} in localized captioning, we adapt it into audio generation.
The video object segmentation-aware control module is fused with the Synchformer's \citep{iashin2024synchformer} Vision Transformer (ViT) \citep{dosovitskiy2020image} based feature extractor, TimeSformer, that is pretrained contrastively on a sub-clip level with audio.

\subsubsection{Focal Prompt}
To provide detailed information coupled with the global context and segmentation information for the visual feature extractor, work by \cite{lian2025describe} introduces a focal prompt.
The focal prompt consists of two different visual stream inputs.
A global video $V$ and its focal crop $V'$, along with their corresponding spatial segmentation masks $M$ and $M'$.
These masks highlight the sounding object (\textit{e.g.}, instruments) in the video stream and are used to steer the audio generation.
The segmentation mask extraction procedure is detailed in \cref{sec:dataset}.
Unlike \cite{lian2025describe}, who paired vision with language, we use these masks directly with a vision transformer tailored for audio-video synchronization.

The cropping is done based on the mask.
We crop the original video so that the masked area is visible throughout the video, but enforce a minimum size of $48\times48$ pixels \citep{lian2025describe}.

\subsubsection{Localized Vision Backbone with Temporal Mask Embedding}
Our approach builds on work by \cite{lian2025describe}, extending it to video-based multimodal generation, where our goal is to extract temporally-aware and localized visual features to control an audio generation process.
To process the focal prompt, we first embed visual and mask streams into a shared spatiotemporal representation.
Specifically, both video and mask streams are reshaped to $224\times224$ spatial size and passed through respective 3D patch embedding layers, $E_V$ and $E_M$.
We apply learnable positional encodings $P \in \mathbb{R}^{T\times197\times768}$, where $T$ is the temporal dimension, 1568 is the sequence dimension, and 786 is the channel dimension.
Positional embeddings are critical to inject temporal ordering and spatial locality and to enable the transformer to learn meaningful correspondences across frames.
The resulting embeddings are:
\begin{equation}
\begin{split}
    & \mathbf{x} = E_V(V) + E_M(M) + P, \\ & \mathbf{x'} = E_V(V') + E_M(M') + P.
\end{split}
\end{equation}

Here, $\mathbf{x}\in \mathbb{R}^{T\times1568\times768}$ and $\mathbf{x'}\in \mathbb{R}^{T\times1568\times768}$ represent the embedded global and focal inputs, respectively.
The mask embedding layer $E_M$ is initialized to output zeros, preventing early-stage training instabilities from disrupting the backbone.
This design follows the initialization strategy in DAM \citep{lian2025describe}.

Next, we extract visual features at two scales.
The global stream $\mathbf{x}$ is processed by a global feature extractor $F_G$, while the focal stream $\mathbf{x'}$ is passed through a regional feature extractor $F_R$ that shares self-attention weights with $F_G$.
This weight sharing encourages alignment and reuse of representations across global and local views:
\begin{equation}
F_{syn} = F_G(\mathbf{x}), \quad F_{syn}' = F_R(\mathbf{x'}, F_{syn}).
\end{equation}

To allow focal features to condition on the global visual context, we adopt gated cross-attention adapters \citep{alayrac2022flamingo, li2022blip}.
These modules are inserted after the self-attention and feedforward layers in each Synchformer block, enabling fine-grained integration of focal and global information. Each block in $F_R$ is updated as follows:
\begin{equation}
\begin{split}
    \mathbf{h}^{(l)'} =~& \mathbf{h}^{(l)} + \tanh(\gamma^{(l)}) \\& \times \text{CrossAttn}(\mathbf{h}^{(l)}, F_{syn}), \\
    \mathbf{h}^{(l)}_{\text{Adapter}} =~& \mathbf{h}^{(l)'} + \tanh(\beta^{(l)}) \times \text{FFN}(\mathbf{h}^{(l)'}).
\end{split}
\end{equation}

Here, $\mathbf{h}^{(l)}$ is the output of the l-th self-attention block in $F_R$, $\gamma^{(l)}$ and $\beta^{(l)}$ are learnable scale parameters initialized to zero to suppress noisy gradients at early training stages.
$\mathbf{h}^{(l)}_{\text{Adapter}}$ is used in place of $\mathbf{h}^{(l)}$ in the preceding Transformer block.
The cross-attention enables focal tokens to selectively attend to relevant global features from $F_{syn}$, improving localization and temporal consistency in the fused visual representation.

The final fused representation $F_{syn}'$ is used to condition the DiT-based audio generator.
By processing video and mask inputs in both global and localized views, we are able to capture fine-grained spatial cues and their temporal dynamics, essential for generating semantically aligned audio.

\begin{figure*}[ht]
    \centering
    \includegraphics[width=1.0\linewidth]{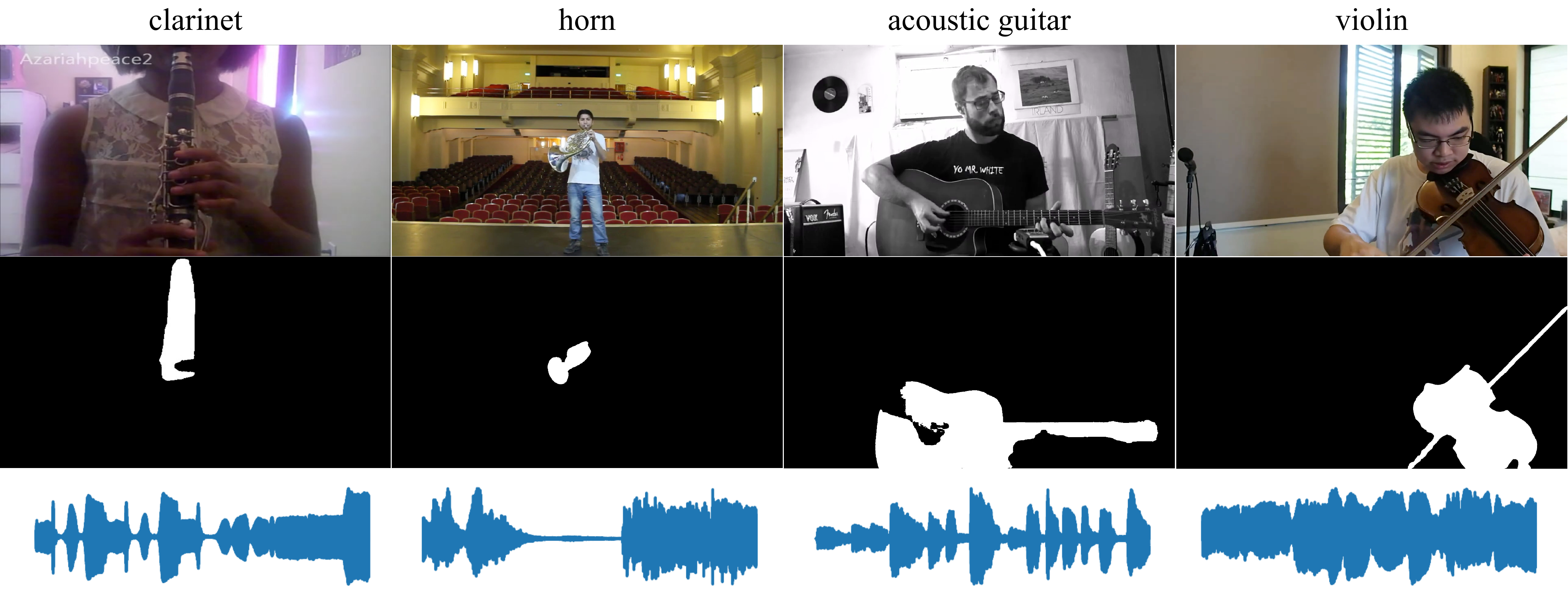}
    \caption{Samples from \datasetName. The top row indicates the musical instrument label. The second row displays the first frames of the video and the corresponding mask stream. Last row displays the audio associated with the sample.}
    \label{fig:sms_sample}
\end{figure*}

\subsubsection{Audio Generation}
Fused visual features are used to condition the audio generation process of MMAudio \citep{chengTamingMultimodalJoint2024} instead of the Synchformer \citep{iashin2024synchformer} features.
Otherwise, the conditioning is kept similar.
Adapting the task formulation described in \cref{ssec:task}, we get $A=G_M(F'_{syn}, F_v, F_t, \mathbf{a})$, ($G_M$ is the MMAudio model, $F'_{syn}$ is the fused visual features, $F_v$ is the visual CLIP \citep{radford2021learning} features, $F_t$ is the textual features, and $\mathbf{a}$ is the noisy audio latent).
For details of the extraction of $F_v$ and $F_t$, please refer to \cite{chengTamingMultimodalJoint2024}.

To further enhance \modelName~performance, we fine-tune the generative model by using Low-Rank Adaptation (LoRA) \citep{hu2021lora} during training. Specifically, we add low-rank matrices for query and value projections of the DiT attention blocks associated with the segmentation-aware visual features.

\section{\datasetName\space Dataset}
\label{sec:dataset}

To facilitate the training of a video object segmentation-aware generative audio model, we propose \datasetName.
We draw inspiration from audio-visual segmentation datasets, where data consists of single and multi-source videos \citep{zhou2022audio, zhou2023audio}.
We hypothesize that by training our model using single-source videos, accompanied by the sounding object masks, our model learns to use the segmentation information for the audio generation.
When a multi-source scene is introduced during test time, the model is capable of generating sound for the segmented object.
Our data pipeline draws inspiration from VGGSound \citep{chen2020vggsound}.

\subsection{Data Sourcing}
The training data consists of video clips of people playing a single instrument and segmentation masks of the instruments.
The raw videos are gathered across multiple datasets.
For the training and validation data, we combine solo performance videos from MUSIC21 \citep{zhao2018sound, zhao2019sound}, AVSBench \citep{zhou2022audio, zhou2023audio}, and Solos \citep{montesinos2020solos}.
For testing, we use the URMP dataset \citep{li2018creating}.
It comprises several multi-instrument musical pieces assembled from separately recorded performances of individual tracks.
Meaning that the final audio is a combination of separate audio recordings.
Thus, each instrument in a multi-source scenario has a separately recorded audio, which is used as the ground truth in video object segmentation-aware audio generation.

\subsection{Visual Verification}
We verify that the target instrument is visually present in the video.
First, we split the video into scenes based on abrupt changes, then clip each scene to avoid transitions that could confuse the segmentation model.
This helps SAM2 \citep{raviSAM2Segment2024} maintain consistent tracking of the object across frames.
Next, we check for the target object in each scene.
We sample frames at 2 FPS and classify them using a pretrained CNN-based model \citep{jocher2022ultralytics} trained on ImageNet \citep{deng2009imagenet}.
If the target appears in the top 5 predictions for a frame, we mark it as present.
Because our target classes don’t always match ImageNet labels directly, we use semantic similarity.
We embed both our $N$ audio classes (depends on the dataset) and the 1000 ImageNet classes using MPNet \citep{song2020mpnet}, then compute cosine similarities to find the top 5 closest ImageNet labels for each audio class.
These act as the visual signatures.
A sample is accepted if at least one of these matching labels appears in the top 5 predictions with a confidence above 0.2.

\subsection{Auditory Verification}
For each visually verified scene, we window the audio to match the temporal dimension of the visual verification process.
For audio classification, we utilize Audio Spectrogram Transformer (AST) \citep{gong2021ast}.
We follow the same procedure as in visual verification to compute the label similarities between the AST classifier and our audio classes.
If the target object is present (or silence is detected) in the top 5 predictions for an audio window, we classify the object as present.
We include silence to support learning of natural pauses during instrument play.

\subsection{Clipping Raw Videos}
Given the scenes, we use the presence information to clip the raw videos into 5 second sequences.
We require that the object be detected within the visual and auditory streams throughout the clip.

\subsection{Mask Generation}
For train data, we utilize the GroundedSAM2 framework \citep{ren2024grounded, raviSAM2Segment2024, kirillov2023segany} and obtain the initial mask by prompting the Florence-2 \citep{xiao2023florence} foundation model with the instrument label.
Florence-2 produces the initial segmentation mask based on a textual prompt, and SAM2 propagates the mask throughout the clip.
Florence-2 was chosen based on its performance on a small manually verified data subset compared with GroundingDINO versions 1.5 and 1.6 \citep{liuGroundingDINOMarrying2024, ren2024grounding}.
For test data, we manually provide location coordinates of the target objects to prompt SAM2.
Manual prompting yielded more coherent masks compared to using a grounding model, but limited resources prevented us from manually annotating target objects in the training data.

Finally, \datasetName\space consists of 5\,395 training, 665 validation, and 745 test 5-second samples spanning over 25 different musical instruments. Every video is accompanied by segmentation masks that have the same number of frames as the video and the label of the segmented object. The frame rate of all samples is 25 FPS, and the audio sample rate is 44.1 kHz.

\section{Experiments}
\subsection{Implementation Details}
We train and evaluate our method using \datasetName\space (\cref{sec:dataset}).
Training data consists of solo musical instrument performance videos, instrument text labels, and segmentation masks.
During training, we drop the textual label with a probability of 50\% to facilitate learning of the segmentation information.
Evaluation data consists of multi-instrument performance videos where the audio is from the segmented instrument, instrument text labels, and segmentation masks.
Following \citep{iashin2022sparse, chengTamingMultimodalJoint2024}, we use H.264 and AAC video and audio encodings, resampled to 25 FPS and 44.1 kHz.
The audio length is set to five seconds.

For our pretrained model, we utilize MMAudio \citep{chengTamingMultimodalJoint2024}.
We train the video object segmentation-aware control module on top of the pretrained model and compare our method against the base model.
We use a learning rate of $1\times10^{-4}$, and AdamW optimizer \citep{loshchilov2017decoupled} with $\beta=[0.9, 0.95]$.
Other training parameters are initialized following \cite{chengTamingMultimodalJoint2024}.
We train on 4 NVIDIA A100 40GB GPUs for $\sim$40 epochs until convergence.
We also experiment with LoRA \citep{hu2021lora} fine-tuning of the query and value projections of the DiT layers associated with the segmentation-aware features.
We use LoRA rank of 16 and set $\alpha=32$.
During testing, we utilize classifier-free guidance \citep{ho2022classifier} with the scale of 7.0.
We use a scale of 4.5 for the base model, as it yields the best performance.

\subsection{Evaluation Metrics}
For fair comparison, we utilize the same evaluation pipeline as described in MMAudio \citep{chengTamingMultimodalJoint2024}.
Quality is evaluated over four different aspects: distribution matching, audio quality, semantic alignment, and temporal alignment.

\subsubsection{Distribution Matching}
We compute Fréchet Distance (FD) and Kullback-Leibler Distance (KL) between generated and ground truth samples.
FD is calculated using VGGish \citep{gemmeke2017audio} (FD$_\text{VGG}$), PANNs (FD$_\text{PANNs}$) \citep{kong2020panns}, and PaSST (FD$_\text{PaSST}$) \citep{koutini2021efficient} embeddings.
KL is calculated using PANN (KL$_\text{PANNs}$) and PaSST (KL$_\text{PaSST}$) embeddings.

Distribution matching in audio quality evaluation assesses how closely the feature distribution of generated audio matches that of ground-truth audio, measured in the embedding space of a pretrained audio classifier.
By measuring the distances of these representations, the generated audio is ensured to be realistic and diverse.
Distribution matching-based metrics are also often utilized in image generation \citep{heusel2017gans}.

\subsubsection{Audio Quality}
We utilize PANNs to calculate Inception Score (IS) \citep{salimans2016improved}.
IS does not compare the generated sample to the ground truth.
It is a metric of objective quality and diversity.

IS was originally adapted for image generation models \citep{salimans2016improved}.
It measures generative model quality by assessing realism (\textit{i.e.} clear classes) and diversity (\textit{i.e.} many classes) using a pre-trained classifier.
In audio quality evaluation, the original IS metric is adapted for spectrograms.

\subsubsection{Semantic Alignment}
We utilize ImageBind \citep{girdhar2023imagebind} to calculate a similarity score (IB-score) between the video and generated audio \citep{viertolaTemporallyAlignedAudio2024}.
ImageBind can embed data from multiple modalities into the same semantically meaningful embedding space.
IB-score is a cosine distance between the audio and video embeddings, and it measures the semantic similarity of the condition (\textit{i.e.} video) and the generated audio.
The ground truth video is cropped to primarily show the segmented instrument and its player.

\subsubsection{Temporal Alignment}
We use Synchformer \citep{iashin2024synchformer} to compute the average of absolute offset predictions between audio and video (DeSync) \citep{viertolaTemporallyAlignedAudio2024}.
Synchformer is able to detect and classify the temporal offset between video and audio streams.
In other words, Synchformer can be used to determine the temporal synchronization between corresponding input streams.
For human viewers, the temporal synchronization of audio and video is vital for high-quality perception \citep{radiocommunicationrelative}.
The ground truth videos are processed similarly as when calculating the IB-score.

\begin{figure}[t]
    \centering
    \includegraphics[width=\linewidth]{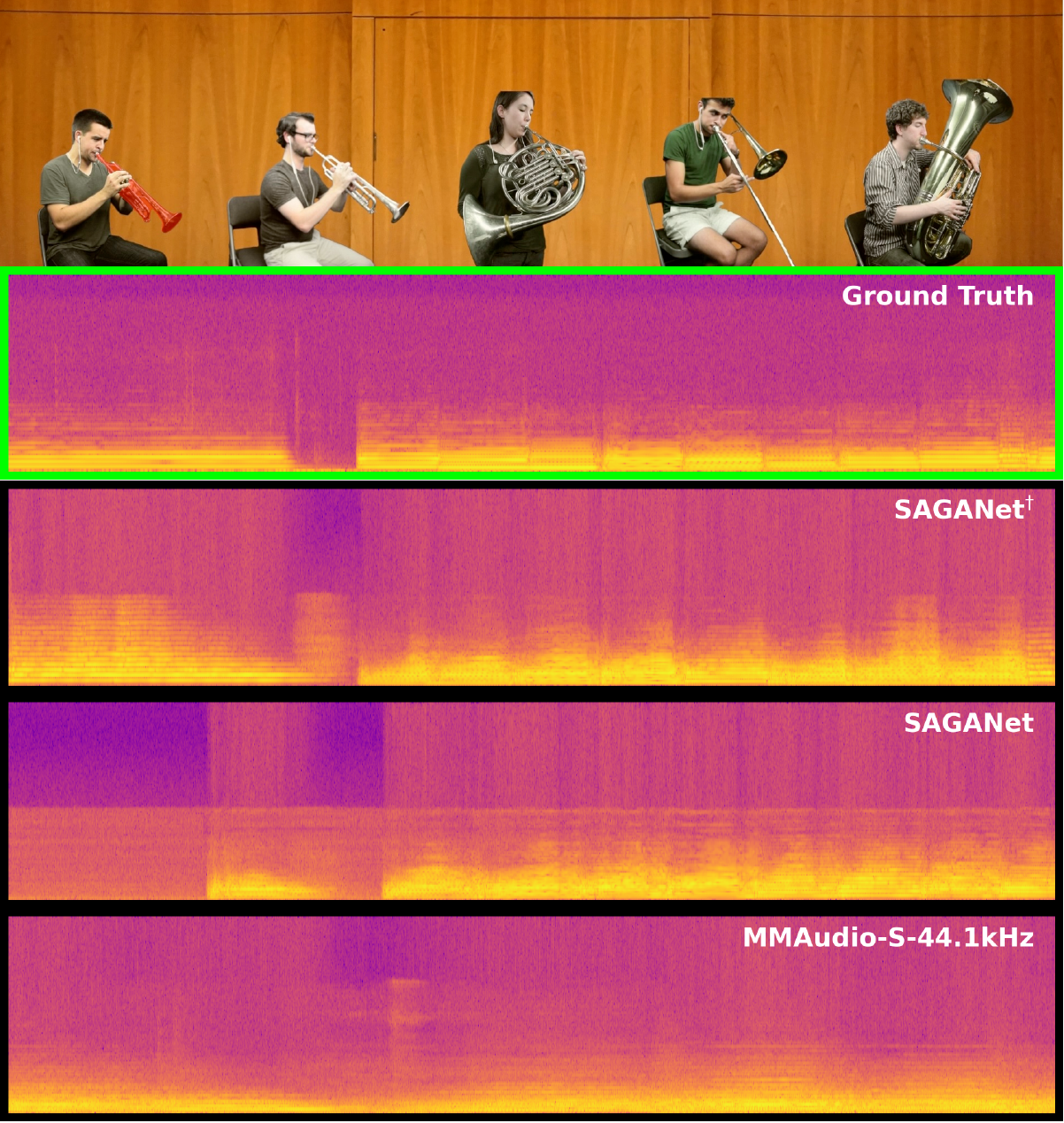}
    \caption{SAGANet generates accurate audio for the target object. Ground truth is a separately recorded audio of the masked object. SAGANet generates matching audio, but MMAudio \citep{chengTamingMultimodalJoint2024} is unable to focus on the target object conditioned solely on video and target instrument label.}
    \label{fig:single-sample}
\end{figure}

\begin{table*}[th]
\caption{\modelName~outperforms the base model. The added video object segmentation-aware control module helps our model to focus on the correct object. MMAudio \citep{chengTamingMultimodalJoint2024} allows guiding the focus only through the textual condition. Results were averaged over 5 samples of \datasetName\space (\cref{sec:dataset}) evaluation data. We utilize the MMAudio-S-44.1kHz variant in our experiments. $\dagger$: Fine-tuned DiT-layers associated with visual segmentation-aware features using LoRA \citep{hu2021lora}.}
\begin{center}
\begin{tabular}{l@{\hspace{.15cm}}|c@{\hspace{.15cm}}c@{\hspace{.15cm}}c@{\hspace{.15cm}}c@{\hspace{.15cm}}c@{\hspace{.15cm}}c@{\hspace{.15cm}}c@{\hspace{.15cm}}c@{\hspace{.15cm}}c}
    \toprule
    & \scriptsize FD$_\text{PaSST}$↓ & \scriptsize FD$_\text{PANNs}$↓ & \scriptsize FD$_\text{VGG}$↓ & \scriptsize KL$_\text{PANNs}$↓ & \scriptsize KL$_\text{PaSST}$↓ & \scriptsize IS↑ & \scriptsize IB-score↑ & \scriptsize DeSync↓ \\
    \midrule
    \scriptsize MMAudio & \multirow{1}{*}{530.60} & \multirow{1}{*}{23.83} & \multirow{1}{*}{\textbf{13.26}} & \multirow{1}{*}{1.17} & \multirow{1}{*}{1.00} & \multirow{1}{*}{2.24} & \multirow{1}{*}{35.94} & \multirow{1}{*}{0.95} & \\
    \scriptsize \modelName & \multirow{1}{*}{364.07} & \multirow{1}{*}{21.31} & \multirow{1}{*}{17.19} & \multirow{1}{*}{0.79} & \multirow{1}{*}{0.59} & \multirow{1}{*}{\textbf{3.34}} & \multirow{1}{*}{39.62} & \multirow{1}{*}{0.44} & \\
    \scriptsize \modelName$^\dagger$ & \multirow{1}{*}{\textbf{330.01}} & \multirow{1}{*}{\textbf{17.88}} & \multirow{1}{*}{19.36} & \multirow{1}{*}{\textbf{0.69}} & \multirow{1}{*}{\textbf{0.56}} & \multirow{1}{*}{3.27} & \multirow{1}{*}{\textbf{41.50}} & \multirow{1}{*}{\textbf{0.30}} & \\
\bottomrule
\end{tabular}
\label{tab:main}
\end{center}
\end{table*}

\begin{table*}[th]
\caption{Average ratings per method from the subjective study. \modelName~outperforms the base model. We utilize the MMAudio-S-44.1kHz variant in our experiments. $\dagger$: Fine-tuned DiT-layers associated with visual segmentation-aware features using LoRA \citep{hu2021lora}. We show the mean $\pm$ std for each method.}
\begin{center}
\begin{tabular}{l@{\hspace{.15cm}}|c@{\hspace{.15cm}}c@{\hspace{.15cm}}c}
    \toprule
    & \scriptsize Audio quality ↑ & \scriptsize Semantic alignment ↑ & \scriptsize Temporal alignment ↑ \\
    \midrule
    \scriptsize MMAudio & 3.60 $\pm$ 1.26 & 2.30 $\pm$ 1.41 & 1.91 $\pm$ 1.21 \\
    \scriptsize \modelName & 3.82 $\pm$ 1.09 & 4.27 $\pm$ 1.14 & \textbf{3.88 $\pm$ 1.18} \\
    \scriptsize \modelName$^\dagger$ & \textbf{4.10 $\pm$ 0.98} & \textbf{4.47 $\pm$ 0.94} & 3.77 $\pm$ 1.25 \\
\bottomrule
\end{tabular}
\label{tab:main_sub}
\end{center}
\end{table*}

\subsection{Results}
\modelName~shows the benefit of video object segmentation-aware control compared to the base model.
Added control is crucial in the scenes where the target object is presented among other sounding instruments.
Note that the evaluation data consists of multi-source videos containing multiple instruments.
Despite being trained solely on single-source samples, our model demonstrates strong generalization to multi-source scenarios, effectively focusing on the target regardless of multiple instruments in the visual input.
Textually describing the target object, accompanied by visual feed, does not provide strong enough guidance for the base model to generate temporally and semantically aligned audio.

\subsubsection{Objective Results}
We report the results in \cref{tab:main}.
We use \datasetName~evaluation set (\cref{sec:dataset}) and average results over 5 generated audios per sample.
During evaluation, MMAudio \citep{chengTamingMultimodalJoint2024} is conditioned on the full frames and with the textual label of the target object.
Still, MMAudio also focuses on other instruments within the scene.
Adding segmentation-aware control guides the model to focus solely on the target object.
This is evident from the strong semantic similarity and temporal alignment achieved by our approach.
Finetuning the DiT layers related to visual segmentation-aware features using LoRA \citep{hu2021lora} improves performance by helping the generative model better adapt to these features.

Our method outperforms the base model on all metrics except FD$_\text{VGG}$.
However, the subjective study in \cref{tab:main_sub} indicates that the audio quality of our approach is higher than that of the base model.
The greatest difference is in the temporal synchronization.
Without the segmentation-aware control module, the model fails to attend to the correct object.
Thus, even though the overall audio quality is sufficient, the temporal alignment is missing.
We present subjective results in \cref{fig:single-sample} and more samples in \cref{app:obj}.

\begin{table*}[ht]
\caption{Ablation study on gated cross-attention (CA). We utilize the MMAudio-S-44.1kHz \citep{chengTamingMultimodalJoint2024} variant in our experiments and fine-tune DiT-layers associated with visual segmentation-aware features using LoRA \citep{hu2021lora}.}
\begin{center}
\begin{tabular}{l@{\hspace{.15cm}}|c@{\hspace{.15cm}}c@{\hspace{.15cm}}c@{\hspace{.15cm}}c@{\hspace{.15cm}}c@{\hspace{.15cm}}c@{\hspace{.15cm}}c@{\hspace{.15cm}}c@{\hspace{.15cm}}c}
    \toprule
    \scriptsize Gated CA & \scriptsize FD$_\text{PaSST}$↓ & \scriptsize FD$_\text{PANNs}$↓ & \scriptsize FD$_\text{VGG}$↓ & \scriptsize KL$_\text{PANNs}$↓ & \scriptsize KL$_\text{PaSST}$↓ & \scriptsize IS↑ & \scriptsize IB-score↑ & \scriptsize DeSync↓ \\
    \midrule
    \scriptsize \tikzxmark & \multirow{1}{*}{396.05} & \multirow{1}{*}{20.41} & \multirow{1}{*}{21.35} & \multirow{1}{*}{0.84} & \multirow{1}{*}{0.65} & \multirow{1}{*}{3.11} & \multirow{1}{*}{41.17} & \multirow{1}{*}{1.04} & \\
    \scriptsize \tikzcmark & \multirow{1}{*}{\textbf{330.01}} & \multirow{1}{*}{\textbf{17.88}} & \multirow{1}{*}{\textbf{19.36}} & \multirow{1}{*}{\textbf{0.69}} & \multirow{1}{*}{\textbf{0.56}} & \multirow{1}{*}{\textbf{3.27}} & \multirow{1}{*}{\textbf{41.50}} & \multirow{1}{*}{\textbf{0.30}} & \\
\bottomrule
\end{tabular}
\label{tab:abl_ca}
\end{center}
\end{table*}

\begin{sidewaystable*}[p]
\caption{Ablation study of different visual prompts. V' and M' refer to using only the detailed crop without global (V, M) visual information. F. Prompt refers to using combined local and global information. Masks (M, M') are embedded and fused with the same strategy as with the proposed method. We utilize the MMAudio-S-44.1kHz \citep{chengTamingMultimodalJoint2024} variant in our experiments. Values are reported as LoRA fine-tuning (no LoRA fine-tuning) \citep{hu2021lora}.}\label{tab:abl}
\centering
\begin{tabular}{l@{\hspace{.15cm}}|c@{\hspace{.15cm}}c@{\hspace{.15cm}}c@{\hspace{.15cm}}c@{\hspace{.15cm}}c@{\hspace{.15cm}}c@{\hspace{.15cm}}c@{\hspace{.15cm}}c}
\toprule
& \scriptsize FD$_\text{PaSST}$↓
& \scriptsize FD$_\text{PANNs}$↓
& \scriptsize FD$_\text{VGG}$↓
& \scriptsize KL$_\text{PANNs}$↓
& \scriptsize KL$_\text{PaSST}$↓
& \scriptsize IS↑
& \scriptsize IB-score↑
& \scriptsize DeSync↓ \\
\midrule

\scriptsize V
& \scriptsize 377.39 {\color{gray}(530.60)}
& \scriptsize 18.29 {\color{gray}(23.83)}
& \scriptsize \textbf{14.94} {\color{gray}(13.26)}
& \scriptsize 0.90 {\color{gray}(1.17)}
& \scriptsize 0.74 {\color{gray}(1.00)}
& \scriptsize 2.80 {\color{gray}(2.24)}
& \scriptsize 39.72 {\color{gray}(35.94)}
& \scriptsize 0.94 {\color{gray}(0.95)} \\

\scriptsize V+M
& \scriptsize 378.68 {\color{gray}(394.46)}
& \scriptsize 18.54 {\color{gray}(19.81)}
& \scriptsize 15.38 {\color{gray}(19.89)}
& \scriptsize 0.91 {\color{gray}(0.84)}
& \scriptsize 0.75 {\color{gray}(0.70)}
& \scriptsize 2.82 {\color{gray}(3.11)}
& \scriptsize 39.82 {\color{gray}(39.25)}
& \scriptsize 0.96 {\color{gray}(0.56)} \\

\scriptsize V'
& \scriptsize 332.05 {\color{gray}(419.30)}
& \scriptsize \textbf{17.34} {\color{gray}(\textbf{17.41})}
& \scriptsize 16.34 {\color{gray}(\textbf{11.88})}
& \scriptsize 0.70 {\color{gray}(0.86)}
& \scriptsize \textbf{0.54} {\color{gray}(0.70)}
& \scriptsize 3.24 {\color{gray}(2.56)}
& \scriptsize 41.37 {\color{gray}(39.19)}
& \scriptsize 0.32 {\color{gray}(\textbf{0.40})} \\

\scriptsize V'+M'
& \scriptsize 384.64 {\color{gray}(\textbf{363.86})}
& \scriptsize 17.65 {\color{gray}(17.74)}
& \scriptsize 16.17 {\color{gray}(17.90)}
& \scriptsize 0.84 {\color{gray}(\textbf{0.74})}
& \scriptsize 0.66 {\color{gray}(0.63)}
& \scriptsize 2.70 {\color{gray}(3.05)}
& \scriptsize 40.05 {\color{gray}(\textbf{39.96})}
& \scriptsize 0.35 {\color{gray}(0.52)} \\

\scriptsize F. Prompt
& \scriptsize \textbf{330.01} {\color{gray}(364.07)}
& \scriptsize 17.88 {\color{gray}(21.31)}
& \scriptsize 19.36 {\color{gray}(17.19)}
& \scriptsize \textbf{0.69} {\color{gray}(0.79)}
& \scriptsize 0.56 {\color{gray}(\textbf{0.59})}
& \scriptsize \textbf{3.27} {\color{gray}(\textbf{3.34})}
& \scriptsize \textbf{41.50} {\color{gray}(39.62)}
& \scriptsize \textbf{0.30} {\color{gray}(0.44)} \\

\bottomrule
\end{tabular}
\end{sidewaystable*}

\subsubsection{Subjective Results}
In addition to objective evaluations, we conduct a subjective user study presented in \cref{tab:main_sub}.
We perform the study for the same models as the objective evaluations.

We select eight evaluation videos from \datasetName~evaluation set.
In total, we collected answers from 21 participants, each one evaluating 24 videos (8 videos $\times$ 3 methods).
The evaluation framework is presented in \cref{app:evalframework}.
Each participant is asked to rate the audio quality in three different aspects using a Likert scale \citep{likert1932technique} (1-5; strongly disagree, disagree, neutral, agree, strongly agree).
Modified from MMAudio \citep{chengTamingMultimodalJoint2024}, we use the following audio quality aspects coupled with the provided instructions:
\begin{enumerate}[\hspace{0.5em}(a)]
    \item The audio has \textbf{high quality}.

    \noindent\emph{Explanation}: An audio is low-quality if it is noisy, unclear, or muffled. In this aspect, ignore visual information and focus on the audio.

    \item The audio is \textbf{semantically aligned} with the segmented target in the video.

    \noindent\emph{Explanation}: An audio is semantically misaligned with the segmented target if the audio is unlikely to occur in the scenario depicted by the video and the segmentation mask, \textit{e.g.}, the sound of a trumpet for a violin.

    \item The audio is \textbf{temporally aligned} with the segmented target in the video.

    \noindent\emph{Explanation}: An audio is temporally misaligned with the segmented target in the video if the audio sounds delayed/advanced compared to the video events, or when audio events happen at the wrong time (\textit{e.g.}, in the video, the drummer hits the drum twice and stops; but in the audio, the sound of the drum keeps occurring).
\end{enumerate}

Subjective results in \cref{tab:main_sub} follow the trend set by objective metrics in \cref{tab:main}.
Our approach exceeds the MMAudio \citep{chengTamingMultimodalJoint2024} across all the measured audio quality aspects.
Temporal and semantic alignment show the greatest improvement over the base method.
Overall audio quality also improves, indicating that the gains are not limited to alignment but extend to subjective audio quality as well.

\subsubsection{Ablation Study}
In \cref{tab:abl}, we analyze the effect of different visual prompts on the generated audio quality.
Masks are embedded and fused in the same way as described in \cref{ssec:arc}.
Also, LoRA is used to fine-tune the same DiT layers across all experiments.
The last row indicates the performance of the proposed model that fuses information across both streams.
Our study shows that both the global and local information are crucial in generating high-quality and aligned audio.

Using only the global visual stream (V) consistently yields the weakest performance in terms of semantic alignment (IB) and temporal synchronization (DeSync) across both LoRA and non-LoRA settings, despite achieving competitive inception scores (IS).
This indicates that while the generated audio may be of reasonable perceptual quality, it lacks relevance to the target object.
Notably, LoRA fine-tuning improves most metrics compared to the non-LoRA counterpart, suggesting more efficient adaptation to the task.

Incorporating segmentation information (V+M) leads to clear improvements compared using only global visual stream, particularly in temporal synchronization and distribution matching in the non-LoRA setting.
However, with LoRA, this configuration results in noticeably higher DeSync.
We attribute this to the adaptation capacity of LoRA: the model can minimize the training objective by refining the general audio generation model (\textit{e.g.}, biasing towards the dominant musical instrument distribution) without sufficiently leveraging the segmentation cues.
As a result, it fails to properly attend to the segmented object, leading to weaker temporal alignment despite the additional spatial information.

Using only local information (V', V'+M') further improves temporal and semantic alignment with the ground truth.
This effect is consistent across both LoRA and non-LoRA configurations, confirming the importance of detailed spatial cues.
However, this comes with a trade-off: while local information significantly boosts semantic alignment (IB) and synchronization (DeSync), tighter localization can harm the overall quality.

Fusing local and global features (F. Prompt) provides the best overall balance.
Under LoRA, this configuration achieves the strongest semantic alignment and temporal synchronization, as well as competitive audio quality, outperforming all other variants in IB-score and DeSync.
In contrast, the non-LoRA variant achieves the highest inception score, but falls short in alignment and synchronization compared to its LoRA counterpart.
This highlights the importance of the fine-tuning for better cross-modal alignment.

In \cref{tab:abl_ca}, we analyze the effect of Gated Cross-Attention (CA) \citep{lian2025describe} on the generated audio quality.
Here, we utilize the focal prompt but use regular cross-attention \citep{vaswaniAttentionAllYou2023} to fuse global and local features.
By utilizing the conventional CA, the model is able to generate high quality (IS) and semantically relevant (IB-score) audio.
However, the largest difference can be seen in the temporal performance (DeSync).
Model without gated CA is not able to generate temporally matching audio for the target object, \textit{i.e.} it is not able to focus on selected single target.
Gated CA is needed to balance the global and local information.

\section{Conclusion}
Current generative audio models lack precise controllability, which limits their introduction \textit{e.g.}, in video post-production.
To address the gap, we introduced a novel audio synthesis task: Video Object Segmentation-Aware Audio Generation.
It enables precise control over audio synthesis by conditioning the audio synthesis on visual segmentation masks.
For the task, we proposed \modelName, a generative audio model conditioned on text, video, and video object segmentation masks.
Our method incorporates both global and localized visual information using a dedicated control module, allowing the model to focus on specific objects within a video.
Our approach significantly improves semantic relevance and temporal alignment, particularly in complex, multi-source scenes where textual or global visual cues fall short.

Furthermore, we presented \datasetName, a benchmark dataset that supports the development and evaluation of VOS-aware audio generation models.
We show that by training our model with single-source videos, our method can generalize to multi-source samples during test time.
This work lays the foundation for the development of more controllable and user-friendly Foley models.

\section{Future Work}
Future work should address the limitations of the proposed method.
This work addresses only audio generation for musical instruments.
Future work will therefore focus on building more diverse datasets, and training and evaluating the proposed method on them.
Additionally, the current approach focuses on generating audio for a single instrument at a time.
Adapting the model to generate audio for multiple selected objects at a time could further clarify its applicability in real-world Foley tasks.

\bmhead{Acknowledgements}

The work was supported by the Academy of Finland projects 353139 and 362409. We also acknowledge CSC – IT Center for Science, Finland, for computational resources.

\begin{appendices}

\section{Subjective Results}\label{app:obj}

\begin{figure*}[p]
    \centering
    \includegraphics[width=0.8\linewidth]{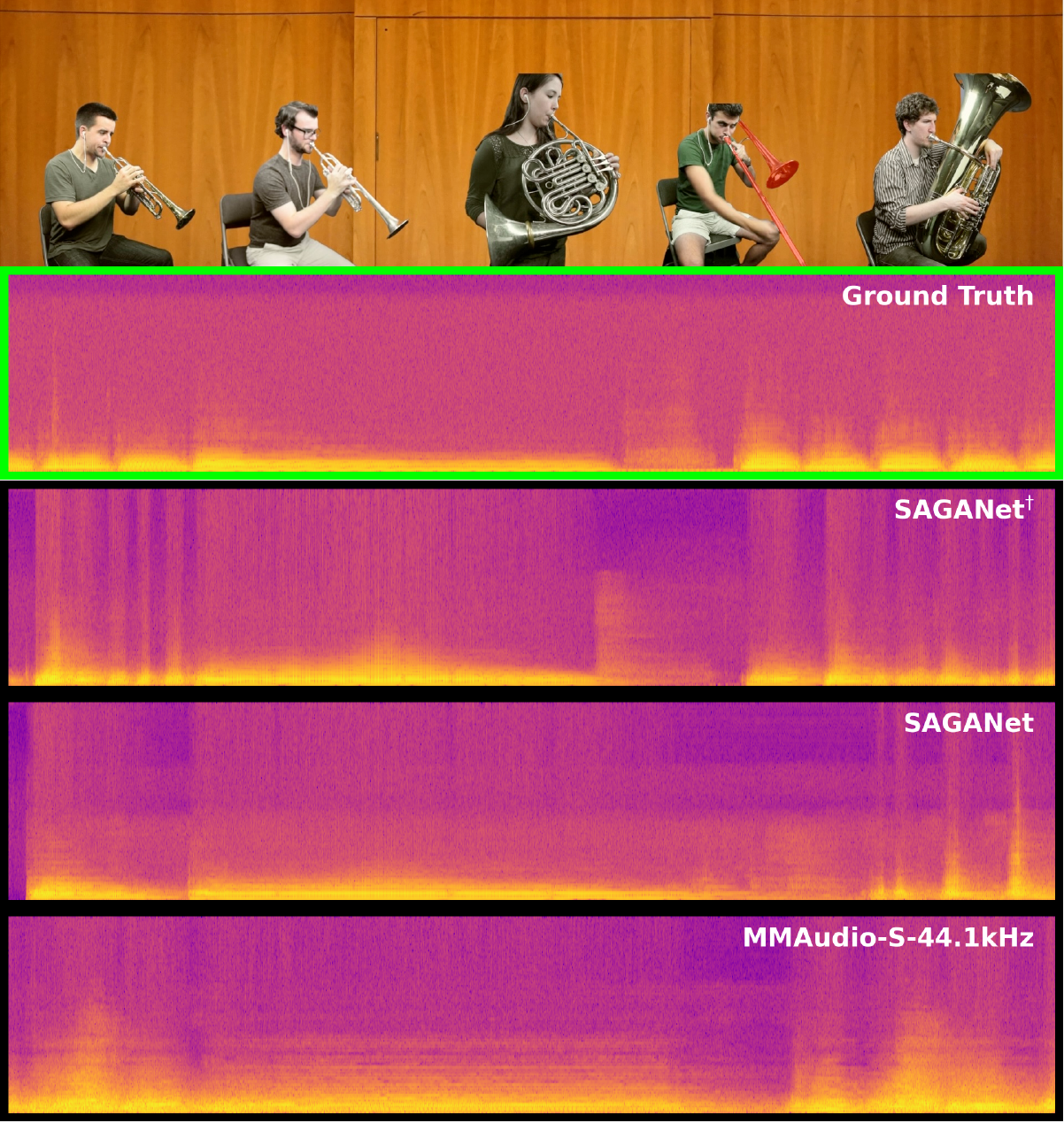}
    \caption{Subjective results on \datasetName~evaluation set. SAGANet generates accurate audio for the target object. Ground truth is a separately recorded audio of the masked object. SAGANet generates matching audio, but MMAudio \citep{chengTamingMultimodalJoint2024} is unable to focus on the target object conditioned solely on video and target instrument label. The target instrument, the trombone second from the right, is highlighted with a red mask.}
    \label{fig:app:sample1}
\end{figure*}

\begin{figure*}[p]
    \centering
    \includegraphics[width=0.8\linewidth]{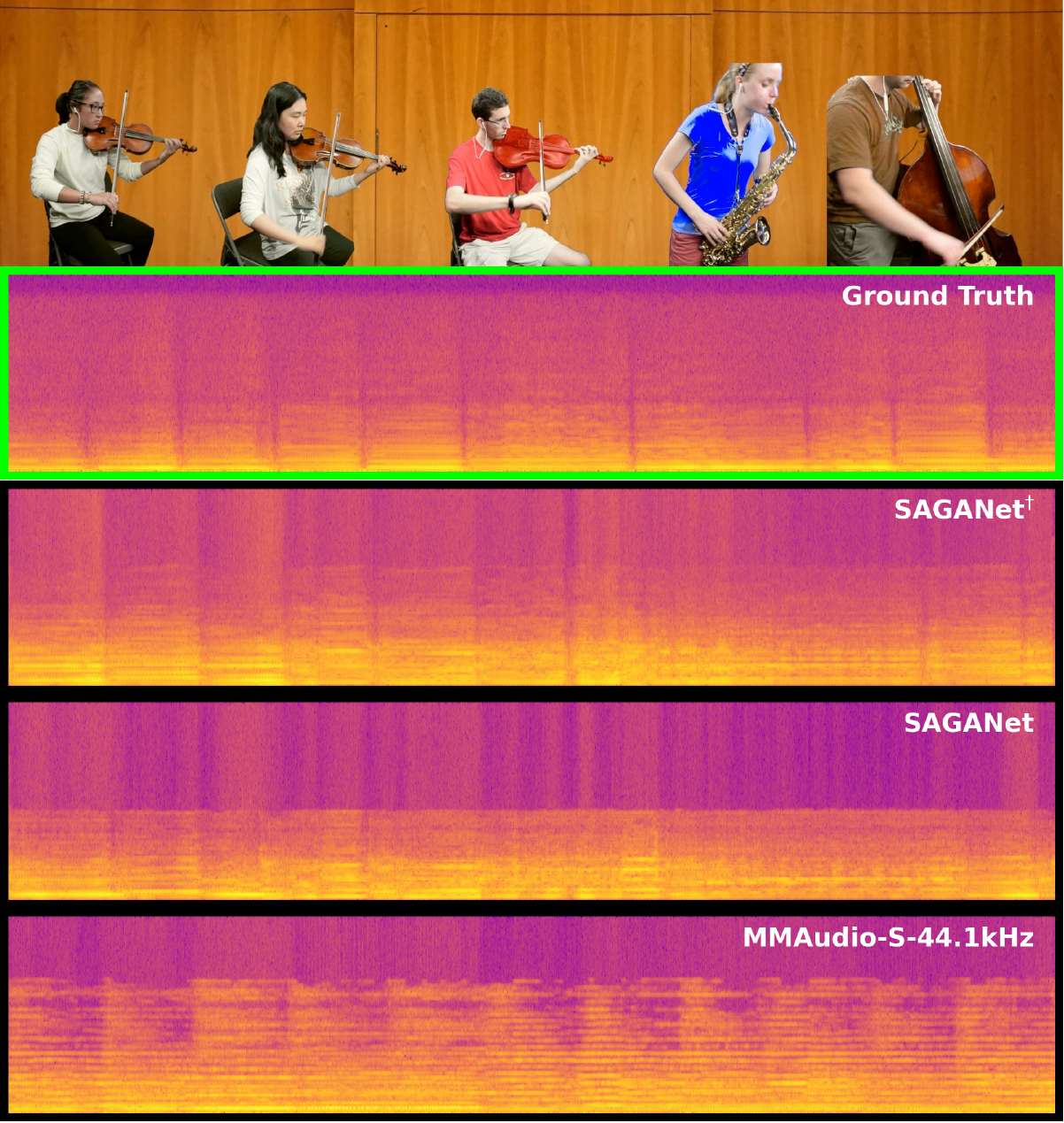}
    \caption{Subjective results on \datasetName~evaluation set. SAGANet generates accurate audio for the target object. Ground truth is a separately recorded audio of the masked object. SAGANet generates matching audio, but MMAudio \citep{chengTamingMultimodalJoint2024} is unable to focus on the target object conditioned solely on video and target instrument label. The target instrument, the violin in the center, is highlighted with a red mask.}
    \label{fig:app:sample2}
\end{figure*}

\begin{figure*}[th]
    \centering
    \includegraphics[width=0.8\linewidth]{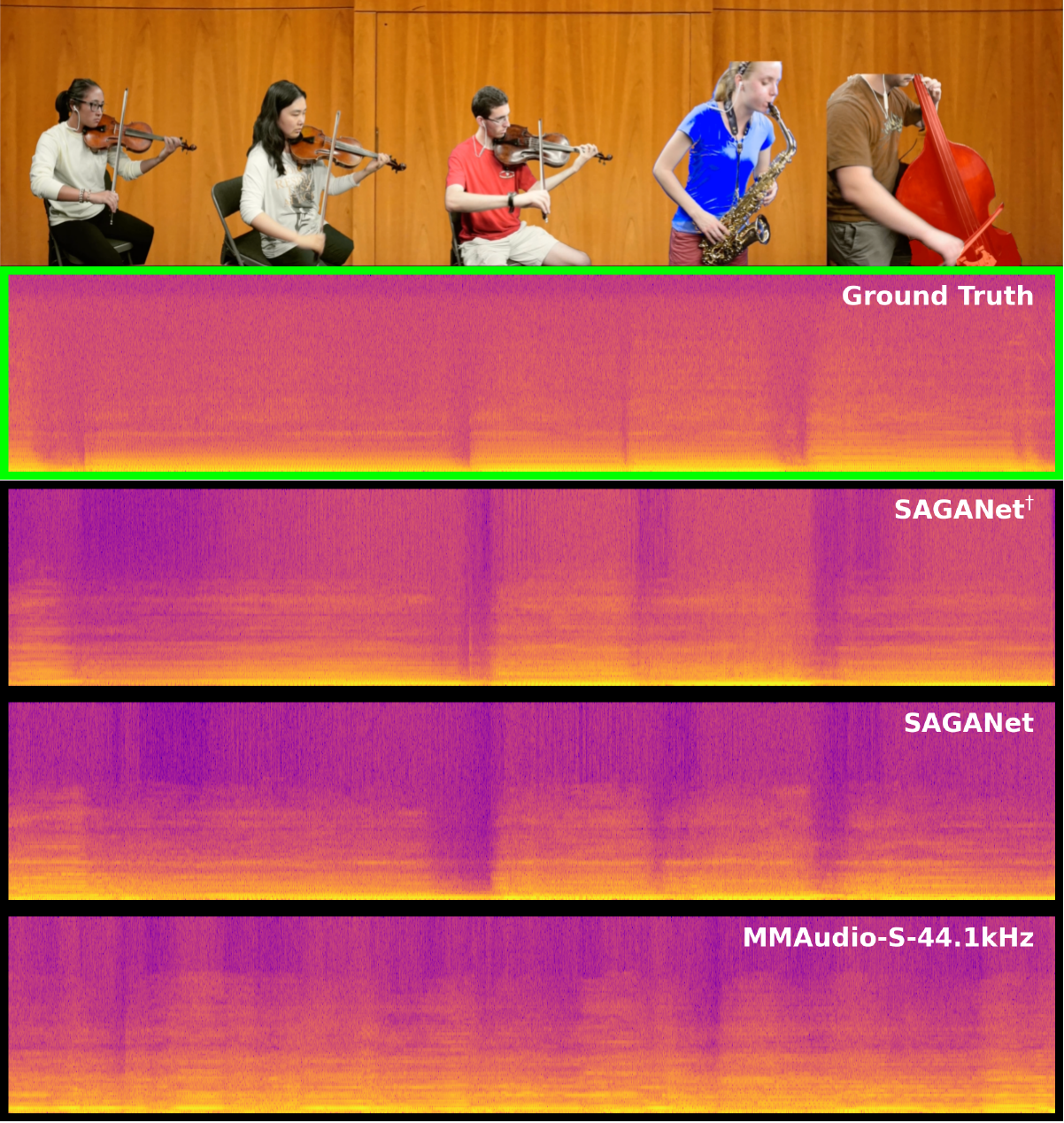}
    \caption{Subjective results on \datasetName~evaluation set. SAGANet generates accurate audio for the target object. Ground truth is a separately recorded audio of the masked object. SAGANet generates matching audio, but MMAudio \citep{chengTamingMultimodalJoint2024} is unable to focus on the target object conditioned solely on video and target instrument label. The target instrument, the double bass on the right, is highlighted with a red mask.}
    \label{fig:app:sample3}
\end{figure*}

\begin{figure*}[p]
    \centering
    \includegraphics[width=0.8\linewidth]{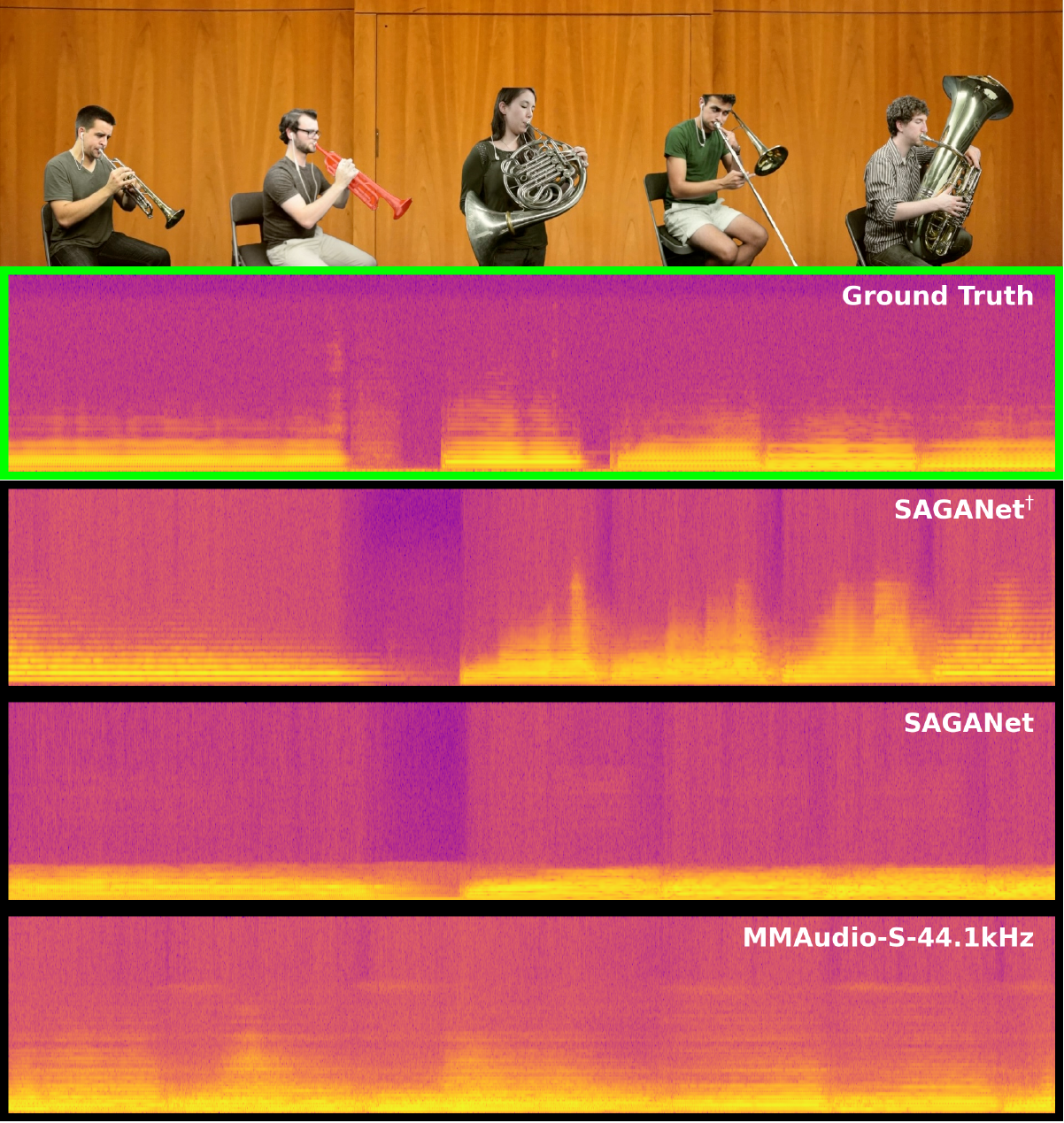}
    \caption{Subjective results on \datasetName~evaluation set. SAGANet generates accurate audio for the target object. Ground truth is a separately recorded audio of the masked object. SAGANet generates matching audio, but MMAudio \citep{chengTamingMultimodalJoint2024} is unable to focus on the target object conditioned solely on video and target instrument label. The target instrument, the trumpet second from the left, is highlighted with a red mask.}
    \label{fig:app:sample4}
\end{figure*}

\Cref{fig:app:sample1,fig:app:sample2,fig:app:sample3,fig:app:sample4} represent subjective samples using \datasetName~evaluation data.
Ground truth is separately recorded audio of the masked object.
SAGANet generates matching audio, but MMAudio \citep{chengTamingMultimodalJoint2024} is unable to focus on the target object conditioned solely on video and target instrument label.
The target instrument is highlighted with a red segmentation mask.

\section{Subjective Evaluation User Interface}\label{app:evalframework}

\begin{figure*}[p]
    \centering
    \includegraphics[width=0.8\linewidth]{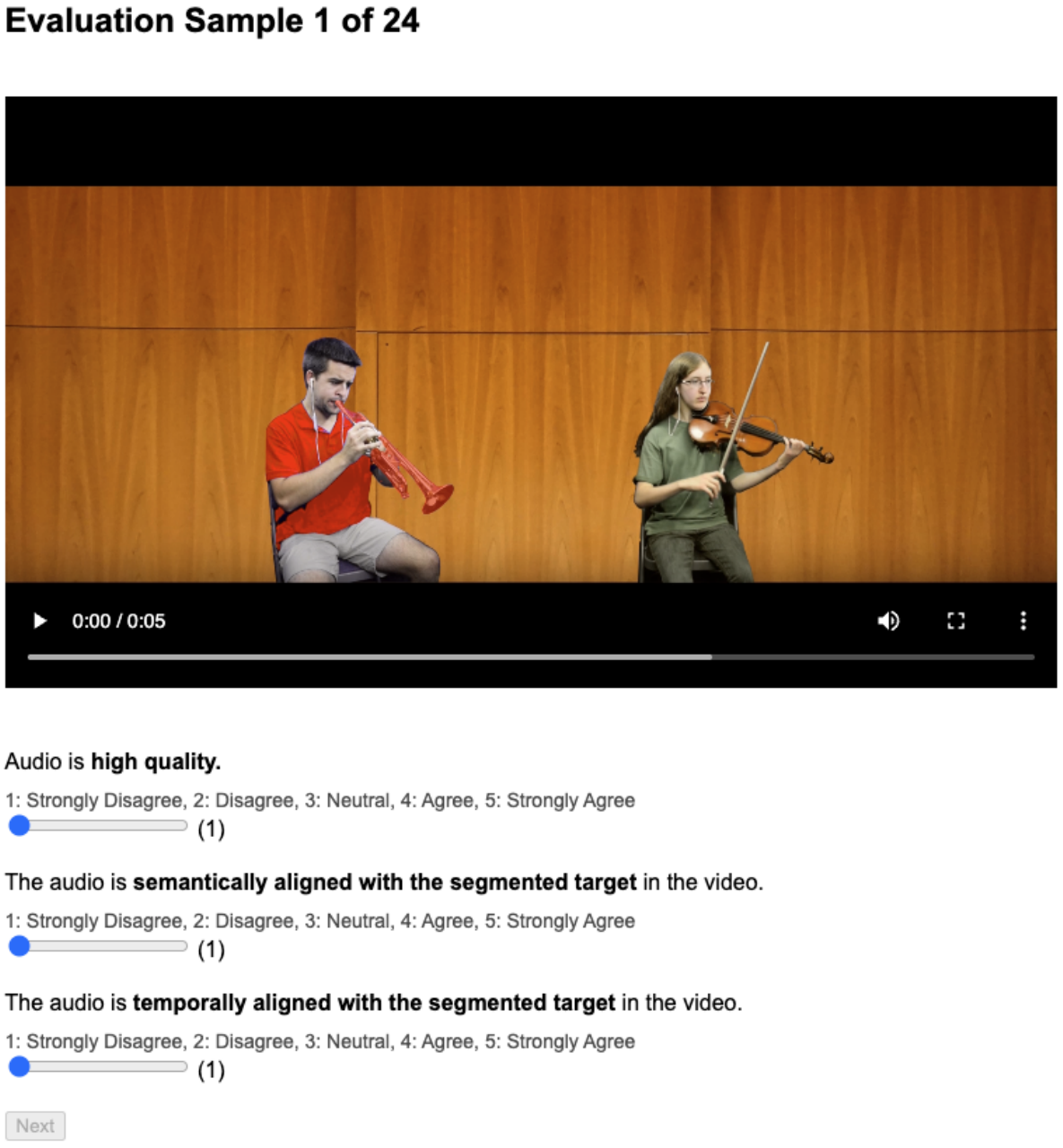}
    \caption{Screenshot of the user interface of the subjective evaluation application. After watching the video, users rate different audio quality aspects of it using the sliders. The segmented object is highlighted with a red segmentation mask throughout the video.}
    \label{fig:app:ui}
\end{figure*}

\Cref{fig:app:ui} presents a screenshot of the user interface of the evaluation application.
First users land to a instruction page, where a video describing the task is presented.
Afterwards the instructions are visible in a written format for the users to revise on.

Participants are shown one video at a time.
Using the three different sliders, they subjectively rate the different quality aspects of the generated audio.
The segmented object is always highlighted with a red mask.
The scores are collected and averaged per generative model to construct a subjective score per model, describing the subjective audio quality of an individual approach.

\end{appendices}

\clearpage
\clearpage
\bibliography{sn-bibliography}

\end{document}